\pdfoutput=1

\documentclass[11pt]{article}

\usepackage{acl}

\usepackage{times}
\usepackage{latexsym}

\usepackage[T1]{fontenc}

\usepackage[utf8]{inputenc}

\usepackage{microtype}

\usepackage{inconsolata}

\usepackage{graphicx}
\usepackage{inconsolata}
\usepackage{graphicx}
\usepackage{amsmath}
\usepackage{adjustbox}
\usepackage{multirow}
\usepackage{svg}
\usepackage[normalem]{ulem}
\useunder{\uline}{\ul}{}
\usepackage{makecell}
\usepackage{tablefootnote}
\usepackage{booktabs}

\usepackage{xcolor}

\definecolor{colorhigh}{HTML}{9900FF}
\definecolor{colorlow}{HTML}{660000}

\title{MOSAIC: Masked Objective with Selective Adaptation for In-domain Contrastive Learning}

\author{Vera Pavlova$^{1,2}$, Mohammed Makhlouf$^{2}$ \\
  $^{1}$burevestnik.ai \quad $^{2}$rttl.ai\\
  $^{1}$\texttt{v@burevestnik.ai,} \quad
  $^{2}$\texttt{v@rttl.ai, mm@rttl.ai}
}
  
\begin{document}
\maketitle
\begin{abstract}
We introduce MOSAIC (\textbf{M}asked \textbf{O}bjective with \textbf{S}elective \textbf{A}daptation for \textbf{I}n-domain \textbf{C}ontrastive learning), a multi-stage framework for domain adaptation of text embedding models that incorporates joint domain-specific masked supervision. Our approach addresses the challenges of adapting large-scale general-domain text embedding models to specialized domains. By jointly optimizing masked language modeling (MLM) and contrastive objectives within a unified training pipeline, our method enables effective learning of domain-relevant representations while preserving the robust semantic discrimination properties of the original model. We empirically validate our approach on both high-resource and low-resource domains, achieving improvements up to 13.4\% in NDCG@10 (Normalized Discounted Cumulative Gain) over strong general-domain baselines. Comprehensive ablation studies further demonstrate the effectiveness of each component, highlighting the importance of balanced joint supervision and staged adaptation. 
\end{abstract}

\section{Introduction}
\label{sec:introduction}
Self-supervised learning has enabled significant progress in natural language processing, with methods like MLM \citep{devlin-etal-2019-bert, liu-etal-2020-roberta, conneau-etal-2020-unsupervised, Sanh-distillation} and contrastive training \citep{reimers-gurevych-2019-sentence, wu2020clear, liu-etal-2021-fast} driving recent developments.
However, these methods are typically explored separately, as effectively combining MLM and contrastive learning remains a significant challenge, since their joint optimization often results in conflicting training signals and suboptimal performance \citep{gao-etal-2021-simcse}.  Nevertheless, unifying these objectives presents an opportunity to leverage the complementary strengths of token-level (MLM) and sentence-level (contrastive) supervision, while also improving the quality of learned representations by mitigating the anisotropy problem (a phenomenon that confines embeddings to a narrow cone-like region in the vector space, thereby limiting their expressiveness) \citep{ethayarajh-2019-contextual, li-etal-2020-sentence, gao-etal-2021-simcse}.
While there have been successful attempts to combine MLM and contrastive objectives for training language models \citep{meng2021coco, chi-etal-2021-infoxlm} and text embeddings \citep{gao-etal-2021-simcse, wu-etal-2022-infocse,  giorgi-etal-2021-declutr}, the majority of the prior work has focused on general-domain data. 

General-domain text embedding models are now widely available, many trained on vast general-domain corpora using a two-stage approach: an initial pre-training phase on massive unlabeled data, followed by supervised fine-tuning \citep{Wang2022TextEB, gte2023, nomic2024, arctic2024}.
The data used for pre-training can exceed half a billion sentence pairs (hundreds of gigabytes of text), resources that are rarely available in specific domains.
Although these general-domain models can perform competitively in specialized areas, their lack of domain-specific knowledge often limits performance. To address this gap, we propose domain adaptation of pre-trained embedding models that leverage their ability to distinguish between similar and dissimilar pairs and transfer it to a domain-specific embedding model.

Previous research on language model adaptation highlights the importance of domain-specific vocabulary for improving results on downstream tasks \citep{beltagy-etal-2019-scibert, Gu-PubMedBERT}. However, simply adding domain-specific vocabulary and continuing MLM training degrades the contrastive properties of the learned representations, since the encoder loses its desirable characteristics under the token prediction objective \citep{wu-etal-2022-infocse}. 
On the other hand, adding new tokens and continuing only with the contrastive objective provides insufficient training signals to update new domain tokens, as the embedding matrix receives diluted signals due to the pooling functions applied to generate text embeddings. 

This dilemma motivates our approach of using a joint objective to enable both token-level and sentence-level supervision, thus benefiting from both worlds and enhancing domain adaptation for both the encoder and the embedding matrix during training. 
Building on a mutual information maximization perspective \citep{Hjelm2018LearningDR, Bachman2019LearningRB, Kong2019AMI, Chen2020ASF, chi-etal-2021-infoxlm}, which demonstrates that these objectives are aligned rather than contradictory, operating at different levels of language granularity, we leverage the joint optimization of MLM and contrastive objectives.
Though these objectives are theoretically aligned, a key challenge in joint training arises from the dominance of the MLM loss and more frequent token-level supervision, which can overwhelm the joint objective and hinder balanced optimization. 
We mitigate this issue by directing the MLM signal toward domain-relevant tokens, rather than separating encoders ~\citep{wu-etal-2022-infocse}, which limits the flow of informative token-level supervision into sentence-level embeddings.
Importantly, MOSAIC builds on a very simple prerequisite: vocabulary expansion. The method only requires adding new domain tokens to the model’s tokenizer, which is a cheap and practical step that avoids complex architectural changes or large-scale retraining. This simplicity makes MOSAIC broadly applicable across domains.

\begin{figure*}[t]
  \includegraphics[width=\textwidth]{{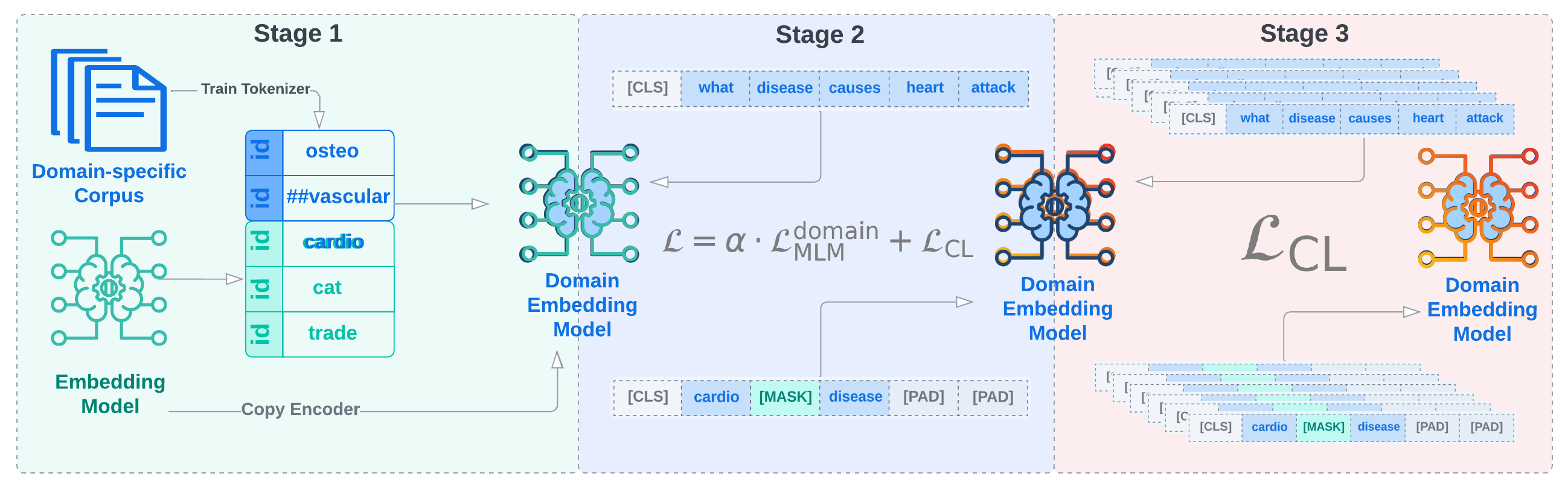}}
  \caption{The MOSAIC pipeline: a multi-stage framework for domain adaptation of text embedding models.}
  \label{fig:approach}
\end{figure*}

To thoroughly evaluate our method, we apply it to both high-resource and low-resource domains.
Most domain-adaptation research focuses on high-resource and medium-resource domains, which is valuable for benchmarking, comparison with strong baselines, and conducting ablation studies.
Yet this focus restricts the generalizability of adaptation methods to truly low-resource domains, which are common in real-world applications. Such domains often face acute data scarcity, making robust adaptation methods essential for ensuring equitable access to state-of-the-art language technologies and maximizing the real-world impact of embedding models \citep{pavlova-makhlouf-2025-efficient}.
To demonstrate the robustness and practical value of our approach, we validate it in two domains: the Biomedical domain, which is characterized by high-resource scientific texts, and the Islamic domain, which represents low-resource but culturally significant content. This allows us to test the robustness of our method even when there is very limited in-domain data.

Our main contributions are as follows:
(1) We propose MOSAIC, a novel domain adaptation approach for pretrained text embedding models that jointly optimizes 
MLM and contrastive objectives within a mutual information maximization framework.
(2) We empirically validate our method on both high-resource (biomedical) and low-resource (Islamic) domains, demonstrating substantial gains over strong general-domain baselines.
(3) We conduct comprehensive ablation studies to analyze the contribution of each component and the dynamics of joint objective training.
(4) We release our code and pretrained models to support reproducibility and facilitate future research \footnote{\url{https://github.com/rttl-ai/mosaic}}.

\section{Related Work}
\label{sec:related}
\subsection{Contrastive and MLM Objectives}

Contrastive Predictive Coding (CPC) introduced the InfoNCE loss \citep{Oord-Contrastive}, encouraging representations to align with an anchor while separating from negatives, thereby laying the groundwork for contrastive learning. In parallel, Masked Language Modeling (MLM), introduced with BERT \citep{devlin-etal-2019-bert}, emerged as the standard pretraining objective for encoder-based models.

Prior work has explored joint contrastive and masked objectives in two distinct directions: large-scale language model pretraining and general-purpose text embedding models. 
COCO-LM integrates contrastive learning into a pretraining pipeline for transformer language models \citep{Vaswani-transformers, meng2021coco}, replacing BERT's Next Sentence Prediction (NSP) objective with a more effective contrastive signal by pairing corrupted and truncated versions of a sentence. The model jointly learns to align these pairs via contrastive loss and to correct the corruption through token-level denoising. COCO-LM demonstrated consistent gains on GLUE tasks \citep{wang-etal-2018-glue}, showing that contrastive objectives outperform NSP for general-purpose pretraining. 
InfoXLM reframes masked language modeling as a contrastive prediction task, formulating it with the InfoNCE loss~\citep{chi-etal-2021-infoxlm} and applying it to multilingual language model pretraining. Combined with a sentence-level cross-lingual contrastive objective, this joint training enables InfoXLM to achieve state-of-the-art results on cross-lingual understanding and retrieval benchmarks.

For training text embeddings, DeCLUTR explicitly combines MLM with contrastive training for text embeddings \citep{giorgi-etal-2021-declutr}. They construct positive pairs from contiguous spans of the same document, apply BERT-style masking to the anchor span, and train jointly through a single encoder. DeCLUTR was evaluated using SentEval, which largely consists of classification tasks along with some STS and NLI tasks.
The authors of InfoCSE introduced a sophisticated framework \citep{wu-etal-2022-infocse} and, rather than combining MLM and contrastive loss on the same encoder output, employed an auxiliary lightweight encoder. While InfoCSE demonstrated gains on STS benchmarks, its architectural separation constrains the transfer of token-level supervision into sentence-level embeddings, limiting the potential benefits of joint training. 

Distinct from these methods, our work introduces a joint MLM-contrastive objective explicitly designed for domain adaptation of existing strong text embedding models,  which, to the best of our knowledge, represents a novel and previously unexplored direction.

\subsection {Domain Adaptation}
Domain adaptation is most commonly performed at the language modeling stage, where general-purpose models undergo continued pre-training on in-domain corpora \citep{Lee-2019, alsentzer-etal-2019-publicly}.  Such approaches typically suffer from the absence of domain-specific vocabulary, which often necessitates training from scratch \citep{beltagy-etal-2019-scibert, Gu-PubMedBERT}. To avoid these GPU-heavy methods, recent work has explored lightweight domain adaptation by introducing new domain vocabulary to already well-trained models, thereby expediting the pre-training process \citep{Poerner-etal-2020-inexpensive, sachidananda-etal-2021-efficient, pavlova-makhlouf-2023-bioptimus, pavlova-2023-leveraging}.

For text embedding models that have already undergone contrastive training, prior efforts have either relied on small adapters \citep{huang-etal-2023-adasent, schopf-etal-2023-efficient} which often suffer from limited expressiveness, or on data-driven methods such as augmentation, denoising objectives, or generative pseudo-labeling \citep{thakur-etal-2021-augmented, wang-etal-2021-tsdae-using, wang-etal-2022-gpl}. 
In contrast, we introduce a full model-driven approach to domain adaptation that operates after language model pretraining. By modifying the training objective itself, our method provides a more expressive and principled mechanism for adapting text embeddings.

\section{Multi-stage Contrastive Learning with Domain-Specific Masked Supervision}
\subsection{Augmenting Contrastive Models with Domain-Specific Vocabulary (MOSAIC-Stage1)}

To leverage the robust encoder learned during contrastive pretraining, we reuse both the encoder and the original embedding matrix. However, to accommodate a word distribution shift from a general domain vocabulary to a new domain vocabulary, we augment the model with new domain-specific tokens (see Figure ~\ref{fig:approach}). The size of the augmented vocabulary in MOSAIC is neither manually selected nor treated as a hyperparameter. Instead, it is a direct and deterministic function of the underlying domain corpus. Given a fixed base tokenizer and a fixed domain corpus, we follow a two-step procedure: 

\textbf{Domain-Specific Tokenizer Training.} We begin by training a new tokenizer on a large domain-specific corpus to identify vocabulary units that capture relevant terminology.

\textbf{Domain Vocabulary Augmentation.} We then identify domain-specific tokens that are missing from the original tokenizer used by the contrastive model, and incorporate these into the model’s embedding matrix, initializing their embeddings as the average of their base model subword embeddings. Tokens outside the base tokenizer’s coverage map to \texttt{[UNK]} and are not added. New domain tokens are constructed as concatenations of existing subtokens, so initialization via the mean of subword embeddings is always well-defined.

This design choice is motivated by the fact that contrastive training mainly shapes the encoder. By modifying only the input vocabulary, we retain the original encoder weights from the pretrained contrastive model, preserving its sentence-level discrimination capabilities.

\subsection{Joint Optimization of Contrastive and MLM Objectives (MOSAIC-Stage2)}
Jointly optimizing MLM and contrastive objectives can theoretically combine the benefits of fine-grained token-level supervision from MLM with sentence-level supervision encouraged by contrastive learning.  However, in practice, it is difficult to perform joint optimization on both. Below, we detail the reasoning behind this challenge and propose our approach to balance these objectives effectively.
To motivate our approach, we start with the information-theoretic interpretation of both MLM and contrastive objectives ~\citep{chi-etal-2021-infoxlm}. Both objectives can be viewed as maximizing a mutual information lower bound. Using the InfoNCE formulation from \citet{Oord-Contrastive}, the contrastive objective for context pairs  $c_1$ and $c_2$ can be expressed as:

\begin{equation}
\label{eq:infonce-contrastive}
I(c_1; c_2) \geq \underset{q(\mathcal{N})}{\mathrm{E}} \left[ \log \frac{f_\theta(c_1, c_2)}{\sum_{c' \in \mathcal{N}} f_\theta(c_1, c')} \right] + \log |\mathcal{N}|
\end{equation}
where $f_\theta$ is a scoring function that measures similarity between two contexts $c_1$ and $c_2$ (e.g., via dot product or cosine similarity), $\mathcal{N}$ represents a set of negative contexts and $q(\mathcal{N})$ sampling distribution of negatives.

Similarly, MLM can also be interpreted as maximizing a mutual information lower bound between the context $c_1$ and the masked token $x_1$ with $\mathcal{V}$ being the vocabulary:

\begin{equation}
\label{eq:infonce-bound}
I(c_1; x_1) \geq \underset{q(\mathcal{V})}{\mathrm{E}} \left[ \log \frac{f_\theta(c_1, x_1)}{\sum_{x' \in \mathcal{V}} f_\theta(c_1, x')} \right] + \log |\mathcal{V}|
\end{equation}

The InfoNCE formulation highlights that while two objectives may be aligned, there is a significant imbalance between them. The larger vocabulary size in MLM results in a substantially larger denominator, leading to very low probabilities for the correct token. Consequently, this generates higher loss values and, therefore, larger gradient magnitudes, causing MLM to dominate the training process.
This imbalance leads to stronger and more frequent gradients for MLM. As a result, the model disproportionately optimizes the MLM objective, leaving the contrastive component under-trained. 
To counteract this, we propose restricting MLM to only a subset of the vocabulary — the domain vocabulary, which includes only newly introduced domain-specific tokens \citep{gu-etal-2020-train, sadeq-etal-2022-informask, Belfathi2024LanguageMA}. This reduces the size of the denominator by replacing the full vocabulary \( V_{\text{all}} \) with a smaller domain-specific set \( V_{\text{domain}} \) limiting the masking signal to rare, informative tokens. Rewriting the MLM loss with domain vocabulary gives:

\begin{align}
\label{eq:infonce-domain-mlm}
&I(c_1; x_1) \nonumber \\
\geq &\underset{q(\mathcal{V}_{\mathcal{D}})}{\mathrm{E}} \left[ \log \frac{f_\theta(c_1, x_1)}{\sum_{x' \in \mathcal{V}_{\mathcal{D}}} f_\theta(c_1, x')} \right] + \log |\mathcal{V}_{\mathcal{D}}|
\end{align}

In this variant, the set \( \mathcal{V}_{\mathcal{D}} \) contains only domain-specific vocabulary tokens.
This targeted vocabulary reduction refocuses the MLM objective on domain-critical tokens, providing clearer and less overpowering gradient signals, which align more closely with those of the contrastive objective.

While the InfoNCE form provides theoretical grounding, in practice both MLM and contrastive learning are usually implemented using cross-entropy losses. For the contrastive loss, this takes the form:

\begin{equation}
\label{eq:contrastive-loss}
\mathcal{L}_{\mathrm{CL}} = -\log \frac{\exp\left(\phi(c_1)^\top \phi(c_2)\right)}{\sum_{c' \in \mathcal{N}} \exp\left(\phi(c_1)^\top \phi(c')\right)}
\end{equation}
where $\phi(\cdot)$ is an encoder that maps the input to a dense vector, and $\mathcal{N}$ includes one positive and $|\mathcal{N}| - 1$ negatives. 

Similarly, the domain-focused MLM cross-entropy loss becomes:

\begin{equation}
\label{eq:crossentropy-domain-mlm}
\mathcal{L}_{\text{MLM}}^{\text{domain}} = -\log \frac{\exp\left(\phi(c_1)^\top e(x_1)\right)}{\sum_{x' \in \mathcal{V}_{\mathcal{D}}} \exp\left(\phi(c_1)^\top e(x')\right)}
\end{equation}
Here, \( \phi \) is the shared encoder (same as used in the contrastive loss), \( e \) is the embedding lookup table, and \( \mathcal{V}_{\mathcal{D}} \) is the (domain-constrained) candidate token vocabulary. 

Our final joint loss is expressed as:
\begin{equation}
\label{eq:joint-loss}
\mathcal{L} = \alpha \cdot \mathcal{L}_{\text{MLM}}^{\text{domain}} + \mathcal{L}_{\text{CL}}
\end{equation}
where $\alpha$ is a scalar coefficient used to balance the gradient magnitude \citep{caruana1997multitask, wu-etal-2022-infocse}. This formulation ensures that both objectives contribute to optimizing the shared encoder while mitigating the gradient dominance of MLM.
Ultimately, by limiting the MLM's vocabulary set and calibrating its contribution to the joint objective, our approach effectively integrates the strengths of MLM and contrastive training, resulting in robust and domain-adapted representations. 
It is crucial to highlight that in our design, the inputs to contrastive objectives are provided with mask perturbation, which forces the model to disambiguate which specific tokens distinguish negative documents from positive (see Figure ~\ref{fig:approach}). In this way, MLM acts as a localized supervision signal that highlights the differences and similarities between pairs, particularly in cases where contrastive loss alone may struggle due to mean pooling (or similar functions), which average over token embeddings and blur these distinctions. By applying MLM-guided masking, the model learns to focus on the key differentiating features.

\subsection{Contrastive-Only Training (MOSAIC-Stage3)}
For the third stage, we continue training our model using only the contrastive objective, after the new domain tokens have been introduced and learned. This stage serves as a corrective step, allowing the encoder to recover and reinforce sentence-level discrimination, which may be diluted during joint MLM+contrastive training. By focusing solely on contrastive learning, the model re-aligns its representations to produce robust text embeddings.

\begin{table*}[t]
    \begin{center}
    \small
    \resizebox{0.98\textwidth}{!}{%
    \begin{tabular}{lccccccccc}
    \toprule
    \textbf{Model} & \textbf{BIOSSES} & \textbf{BiorxivC} & \textbf{MedicalQAR} & \textbf{MedrxivP2P} & \textbf{MedrxivS2S} & \textbf{NFCorpus} & \textbf{SciFact} & \textbf{TRECCOVID} & \textbf{Avg.} \\
    \midrule
    \midrule
    \multicolumn{10}{c}{\it{Unsupervised models}} \\
    \midrule
    nomic-embed-text-v1\textsubscript{unsup}   & 87.189  & 38.78  & 68.307   & 34.854  & \textbf{32.521}  & \textbf{35.684}  & 71.982    & \textbf{62.203} & 53.940 \\
    nomic-embed-bio\textsubscript{unsup}      & 87.012  & 36.107          & 66.173   & 30.72            & 28.552          & 34.235          & 73.302    & \textbf{62.203} & 52.788 \\
    MOSAIC-Bio-Stage1        & 78.946  & 33.747          & 63.58    & 30.261           & 25.126          & 26.091          & 67.246    & 57.050 & 47.756 \\
    MOSAIC-Bio-Stage2        & \textbf{88.116}  & 38.101          & 68.677   & 34.536           & 29.882          & 32.217          & 72.535    & 60.763 & 53.853 \\
    MOSAIC-Bio-Stage3        & 88.057  & \textbf{39.31}           & \textbf{70.233}   & \textbf{35.089}           & 30.287          & 34.137          & \textbf{74.710}    & 61.281 & \textbf{54.638} \\
    \midrule
    \multicolumn{10}{c}{\it{Supervised models}} \\
    \midrule
    all-mpnet-base-v2                     & 80.432  & 42.31           & 66.517   & \textbf{39.862}           & \textbf{37.023}          & 33.289          & 65.570    & 51.326 & 52.541 \\
    E5\textsubscript{base} \citep{Wang2022TextEB}                & 85.103  & 37.49           & 68.051   & 34.6347          & 32.0616         & 36.589          & 73.083    & 79.638 & 55.956 \\
    E5-base-peft-biomed      & 85.017  & 38.049          & 64.470   & 33.730           & 32.665          & 33.921          & 70.152    & 44.728 & 50.342 \\
    GTE\textsubscript{base} \citep{gte2023}               & 87.642  & 40.62           & 71.455   & 36.404           & 34.9025 & \textbf{37.897}  & \textbf{76.178}    & 68.783 & 56.738 \\
    BGE\textsubscript{base} \citep{Xiao2023CPackPR}               & 85.533  & -               & -        & -                & -               & 35.539          & 73.258    & 76.447 & 67.944 \\
    text-embedding-ada-002                & 86.351  & -               & -        & -                & -               & 36.972          & 72.746    & 68.474 & 66.636 \\
    nomic-embed-text-v1\textsubscript{super}                   & 86.471  & 41.48           & 66.648   & 37.0082          & 34.3009         & 35.028          & 70.500    & \textbf{79.923} & 56.670 \\
    nomic-embed-bio\textsubscript{super}      & 70.55  & 35.35           & 52.19    & 31.19            & 25.20            & 31.37           & 47.45    & 56.48 & 43.72 \\
    PubMedBERT-embed                      & 87.567  & 37.874          & 68.349   & 36.270           & 29.865          & 37.430          & 72.719    & 65.754 & 54.478 \\
    MOSAIC-Bio\textsubscript{super}                     & \textbf{89.869} & \textbf{42.551} & \textbf{72.378} & 37.865 & 32.631          & 35.571          & 75.875    & 63.546 & \textbf{56.774} \\
    \bottomrule
    \end{tabular}
    }
    \end{center}
    \caption{
        Evaluation of unsupervised and supervised models across biomedical benchmarks. Bold indicates the highest score per column within each group.
    }
    \label{tab:biomed_main}
\end{table*}

\section{Experiments on a High-Resource Domain}
\textbf{Training Data}.
To construct a large-scale biomedical corpus, we parsed the 2025 PubMed snapshot and extracted \textit{(title, abstract)} pairs. When available, metadata such as journal name and keywords were appended to the title to enrich the context. We filtered out non-English entries as well as pairs where either the title or abstract was too short to form a meaningful sentence pair.
To further ensure data quality and minimize false positives, we applied a consistency-based filtering procedure using the \texttt{gte-base} model (see Appendix~\ref{sec:false_positive}). This resulted in approximately 20 million high-quality sentence pairs for use in stages two and three of our approach (MOSAIC-Bio-Stage2 and MOSAIC-Bio-Stage3).

\textbf{Fine-tuning data}. We evaluate our models in a zero-shot setting. To prevent benchmark data leakage (a common concern for text embedding models), we fine-tune on BioASQ Task 9a \citep{tsatsaronis2015bioasq}, which does not overlap with any of the MTEB evaluation datasets. The dataset consists of approximately 16 million PubMed title–abstract pairs annotated with MeSH (Medical Subject Headings); Importantly, this fine-tuning step is not part of the proposed MOSAIC training pipeline. Instead, it corresponds to preparing the final supervised model (MOSAIC-Bio\textsubscript{super}) used for evaluation in supervise setting.

\textbf{Evaluation Data and Metrics}.
We evaluate on the medical subset of the MTEB (Massive Text Embedding Benchmark) \citep{MTEB}, a standardized benchmark for assessing the quality of text embeddings across a diverse set of tasks, such as retrieval, classification, clustering, reranking, semantic textual similarity (STS), and summarization. We use BiorxivClusteringP2P, MedrxivClusteringP2P, and MedrxivClusteringS2S for clustering (V-measure); MedicalQARetrieval, NFCorpus, SciFact, and TRECCOVID for retrieval (NDCG@10); and BIOSSES for STS (Spearman correlation). We report results on BIOSSES in Table~\ref{tab:biomed_main}, but the analysis on the STS task is performed as a part of the ablation Section~\ref {sec:ablation}. 

\textbf{Baselines}.
For unsupervised baselines, we use {nomic-embed-text-v1\textsubscript{unsup}\footnote{\url{https://huggingface.co/nomic-ai/nomic-embed-text-v1-unsupervised}} as our primary baseline representing an unsupervised contrastive embedding model pretrained on general-domain data. We also train this model on the unsupervised training data described above and include the \texttt{nomic-embed-bio} model in the comparison. 
To analyze the impact of each stage on domain adaptation, we evaluate three variants of our MOSAIC pipeline, all initialized from the nomic-embed-text-v1\textsubscript{unsup} checkpoint. MOSAIC-Stage1 augments the model with new domain-specific vocabulary without further pretraining. MOSAIC-Stage2 applies joint training with masked language modeling and contrastive objectives on domain data. MOSAIC-Stage3 continues training with the contrastive objective alone, yielding the final unsupervised MOSAIC model.
For supervised baselines, we select a diverse set of well-established embedding models that report MTEB scores on biomedical datasets, as listed on the official MTEB leaderboard\footnote{\url{https://huggingface.co/spaces/mteb/leaderboard}}.
As noted in Section \ref{sec:related}, most prior efforts on text embedding adaptation have focused on data-driven techniques and do not produce a general-purpose model checkpoint suitable for evaluation or reuse. Moreover, the biomedical embedding landscape remains sparse: to the best of our knowledge, there is no research-based, dedicated biomedical sentence embedding model available on MTEB. To address this gap, we prepare a biomedical text embedding model derived from the state-of-the-art PubMedBERT, which we further train on the domain-specific data described above (PubMedBERT-embed). We additionally include an adapter-based baseline (E5-base-peft-biomed), where LoRA adapters are applied to E5\textsubscript{base} on the same domain corpus

\textbf{Implementation Details}.
We implement the joint MLM and contrastive training on top of the Nomic repository\footnote{\url{https://github.com/nomic-ai/contrastors/tree/main}}. For the purely contrastive stage, we reuse the original implementation from the repository. The model architecture is based on BERT \citep{devlin-etal-2019-bert} with several modifications introduced by the Nomic repo. In MOSAIC pipeline, we first expand the tokenizer by adding approximately 9k new biomedical tokens, and then implement a joint training objective that combines masked language modeling with contrastive learning.
We set the masking rate to 0.15, the MLM loss weighting hyperparameter \(\alpha = 0.3\) throughout the joint training phase. Details of ablation on $\alpha$ and masking rate can be found in Section~\ref{sec:ablation}, and other hyperparameter settings are provided in Appendix~\ref{sec:hyperparameters}. We train stages two and three of the proposed pipeline using only in-batch negatives, and additionally include hard-mined negatives during fine-tuning.

\subsection{Results and Analysis}
Our results demonstrate several important trends regarding domain adaptation for text embeddings (see Table~\ref{tab:biomed_main}). First, among unsupervised baselines, we observe that simply continuing pretraining a general-domain embedding model (\texttt{nomic-embed-text-v1\textsubscript{unsup}}) on in-domain data can lead to reduced performance compared to the original general-domain baseline across most benchmarks (as in the \texttt{nomic-embed-bio} model), suggesting that naive in-domain adaptation may distort learned representations. This issue becomes even more pronounced when augmenting the vocabulary with domain-specific tokens without any retraining (MOSAIC-Bio-Stage1), resulting in a substantial performance drop across all datasets, likely due to embedding mismatch. In contrast, applying a joint MLM+contrastive objective (MOSAIC-Bio-Stage2) recovers and further improves results, while a final contrastive-only training stage (MOSAIC-Bio-Stage3) achieves the highest scores on four benchmarks (BiorxivClusteringP2P, MedicalQARetrieval, MedrxivClusteringP2P, and SciFact), resulting in the best average performance overall with a 2.8\% increase over the general-domain baseline. These results highlight a clear progression across adaptation stages, where naive vocabulary expansion leads to degradation, targeted joint supervision restores model quality, and a final contrastive stage enables robust domain adaptation.

In the analysis of supervised baselines, we observe that using a general-domain embedding model that has undergone supervised training (\texttt{nomic-embed-text-v1\textsubscript{super}}) as the starting checkpoint for domain adaptation, and further training it on biomedical data (\texttt{nomic-embed-bio\textsubscript{super}}), results in lower performance than adapting a model pretrained only in an unsupervised manner on general-domain data (\texttt{nomic-embed-text-v1\textsubscript{unsup}}).

We further find that the adapter-based approach (E5-base-peft-biomed) degrades the performance of E5\textsubscript{base}, placing it behind other supervised baselines across all evaluated tasks. Additionally, tuning a biomedical language model (PubMedBERT) in a contrastive regime on biomedical data (PubMedBERT-embed) does not yield superior performance on any benchmark. This suggests that transferring contrastive knowledge from a strong general-domain embedding model is more effective than training a domain-specific LLM from scratch with a contrastive objective on in-domain data.

We also observe that competing baselines exhibit task-specific strengths: for example, all-mpnet-base-v2 performs particularly well on clustering tasks, while GTE\textsubscript{base} and nomic-embed-text-v1 show strong performance in retrieval. In contrast, MOSAIC-Bio\textsubscript{super} demonstrates consistently strong performance across task types, including semantic textual similarity (BIOSSES), clustering (MedrxivClusteringP2P), and retrieval (MedicalQARetrieval), achieving the highest average score overall.

However, as shown in Table~\ref{tab:biomed_main}, all of our models lag behind on the TRECCOVID dataset \citep{voorhees2021trec}. Inspection of the TRECCOVID queries reveals that, alongside core biomedical and clinical questions, a significant fraction focuses on social or policy aspects of the pandemic (approximately 20\%). Such queries, addressing societal impacts or interventions like school reopening, may fall outside the primary scope of the biomedical corpora used for model adaptation. This mismatch in domain coverage could partly explain the observed underperformance. Moreover, recent large-scale analyses of PubMed using embedding-based atlases have shown that COVID-19 literature forms a uniquely isolated cluster in embedding space, with strong internal topical fragmentation, further challenging biomedical models~\citep{kobak2024landscape}. The qualitative analysis on BIOSSES is provided in the Appendix~\ref{sec:qualitative_analysis}.

\subsection{Ablation Studies}
\label{sec:ablation}

\begin{table}[t]
\centering
\small
\resizebox{1.0\columnwidth}{!}{%
\begin{tabular}{lc}
\hline
\textbf{Model} & \textbf{Score} \\
\hline
MOSAIC-Stage2 $\alpha$=0.1 & 76.032 \\
MOSAIC-Stage2 $\alpha$=0.2 & 81.336 \\
MOSAIC-Stage2 $\alpha$=0.3 & \textbf{88.116} \\
MOSAIC-Stage2 $\alpha$=0.4 & 86.794 \\
MOSAIC-Stage2 $\alpha$=0.5 & 67.708 \\
All-Token MLM, $\alpha$=0.3 & 63.995 \\
All-Token MLM, $\alpha$=0.1 & 66.092 \\
All-Token MLM, $\alpha$=0.005 & 77.768 \\
All-Token MLM, $\alpha$=0.001 & 78.753 \\
MOSAIC-Stage2 (mlm\_prob 0.3) & 49.871 \\
MOSAIC-Stage2 (as the 3d stage) & 70.540 \\
Contrastive only & 84.428 \\
\hline
\end{tabular}
}
\caption{Performance comparison of MOSAIC-Stage2 and ablated variants on BIOSSES.}
\label{tab:bjmc_vs_ablated}
\end{table}

The primary focus of our ablation study is the second stage of the proposed method (joint MLM+contrastive training). Accordingly, all ablation experiments are conducted on the MOSAIC-Bio-Stage2 model.
For ablation, we use BIOSSES~\citep{biosses}, an STS dataset that requires models to capture fine-grained semantic relationships between sentences, beyond what is assessed in standard retrieval or clustering tasks~\citep{cer-etal-2017-semeval}; this enables us to demonstrate the effect of the MLM objective.

\textbf{Alpha hyperparameter}.
We ablate the effect of the MLM loss weight ($\alpha$), which controls the relative contribution of the MLM objective during joint training. 
We systematically explore a range of $\alpha$ values, a hyperparameter whose impact is rarely examined in prior literature.
As shown in Table~\ref{tab:bjmc_vs_ablated}, setting $\alpha = 0.5$ causes the MLM loss to dominate, resulting in a drastic performance drop. At the other extreme, $\alpha = 0.1$ does not sufficiently promote learning of new domain tokens, and $\alpha = 0.2$ yields only modest gains. While $\alpha = 0.4$ remains competitive though slightly suboptimal, the highest performance is achieved at $\alpha = 0.3$, indicating it as the most balanced choice for our joint objective.

\textbf{Effect of Masking Strategy}.  
To evaluate the effectiveness of domain-restricted masked language modeling (MLM), we compared our default approach, which restricts MLM to domain-specific tokens, with an alternative that applies MLM to all vocabulary tokens (All-Tokens MLM), using $\alpha \in \{0.3, 0.1, 0.005, 0.001\}$ (see Table~\ref{tab:bjmc_vs_ablated}).  
In line with the findings of SimCSE \citep{gao-etal-2021-simcse}, who conducted a small experiment in their Appendix on the impact of MLM when combined with a contrastive objective, we observe that even very small values of $\alpha \in \{0.005, 0.001\}$ negatively affect performance when using random masking over the full vocabulary and remain below those of the domain-token strategy.  
This underscores our central finding: the masking mechanism itself—restricting MLM to domain tokens—matters more than fine-grained tuning of $\alpha$. As $\alpha \to 0$, the all-token variant collapses toward contrastive-only training, which negates the intended benefits of joint supervision and still fails to match domain-token masking.

\textbf{No Joint Objective.}
We assess the impact of the joint MLM+contrastive objective by removing the second stage entirely and training solely with the contrastive objective after vocabulary expansion (“Contrastive only”). As shown in Table~\ref{tab:bjmc_vs_ablated}, omitting the MLM stage results in a performance drop from 88.1 to 84.4, indicating that joint training with domain-restricted MLM provides a meaningful boost over contrastive adaptation alone.

\textbf{Masking rate}.
We further ablate the effect of the masking rate by increasing the MLM probability from the default 0.15 to 0.3 during joint training. As shown in Table~\ref{tab:bjmc_vs_ablated}, raising the masking rate leads to a dramatic drop in performance (from 88.1 to 49.9), indicating that excessive masking can overwhelm the contrastive signal and degrade the learned representations.

\textbf{Order of Training Stages}.
Next, we reverse the order of stages 2 and 3 by first performing only contrastive training with a large batch, followed by contrastive training combined with MLM. As shown in Table~\ref{tab:bjmc_vs_ablated}, this results in a noticeable performance drop from 88.116 to 70.548, a decrease of 20\%. This suggests that applying the joint objective to an already strong embedding model can disturb its contrastive capability.

\section{Experiment on a Low-Resource Domain}

\begin{table}[t]
\centering
\small
\resizebox{0.8\columnwidth}{!}{%
\begin{tabular}{lc}
\hline
\textbf{Model} & \textbf{NDCG@10} \\
\hline
Islamic-embed-model & 33.581*\\
MOSAIC-ID-Stage1 & 30.050 \\
MOSAIC-ID-Stage2 & 34.670 \\
MOSAIC-ID-Stage3 & \textbf{36.809} \\
all-mpnet-base-v2 & 31.516** \\
nomic-embed-text-v1 & 32.048** \\
E5\textsubscript{base} & 32.466* \\
GTE\textsubscript{base} & 32.924* \\
BGE\textsubscript{base} & 27.699** \\
\hline
\end{tabular}
}
\caption{NDCG@10 evaluation results on the Islamic dataset. 
* indicates statistical significance at $p<0.1$ and ** at $p<0.05$ (paired t-test vs.\ MOSAIC-ID-Stage3).}
\label{tab:ndcg10_islamic}
\end{table}

\begin{figure}[t]
\centering
  \includegraphics[width=\columnwidth]{{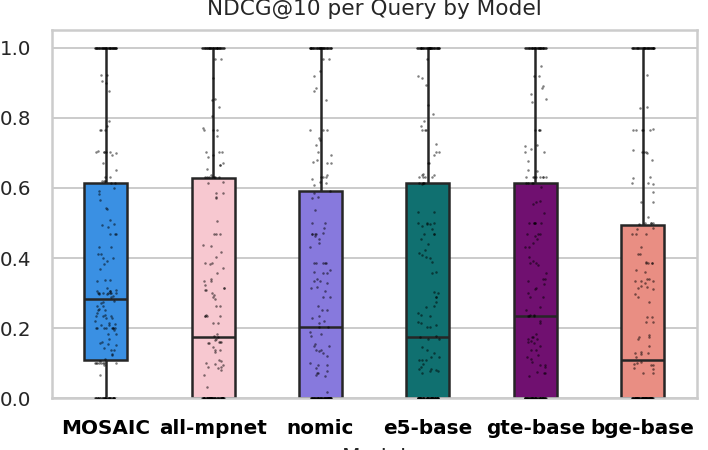}}
  \caption{Boxplot of per-query NDCG@10 scores on the Islamic domain dataset.}
  \label{fig:boxplot_islamic}
\end{figure}

\textbf{Experimental Setup}.
As noted in Section~\ref{sec:introduction}, the Islamic domain is a low-resource area, especially for English-language data. In-domain data suitable for training text embedding models is extremely scarce. To address this, we constructed an in-domain training set by extracting semantically related verse pairs from \textit{Tafseer Ibn Kathir} and applying consistency filtering with the \texttt{gte-base} model, resulting in 7,587 high-quality pairs. Further details on the data construction process are provided in Appendix~\ref{sec:islamic_passages}.

Although the Islamic domain in English is characterized by an extremely limited amount of available training data, it is notable for having a dedicated evaluation dataset—unlike many other low-resource domains. Recent efforts by \citet{malhas2020ayatec} have created a verified high-quality Qur'anic Reading Comprehension Dataset (QRCD), which includes questions frequently asked within the Islamic domain. 
The answers provided are exhaustive, meaning all Qur'anic verses directly responding to the questions have been thoroughly extracted and annotated. To increase the size of the evaluation set, we combine the training and development splits, resulting in a total of 169 queries for testing. Although QRCD is originally in Arabic, we employ verified English translations to enable evaluation in the English language \citep{pavlova-2025-multi}. For retrieval collections, we use the Sahih International English translation.\footnote{\url{https://tanzil.net/trans/}}
We compare our final model with three strong general-domain embedding models using NDCG@10 as the evaluation metric.
The implementation details follow those used for the biomedical model, with the following modifications: we add 3k domain-specific tokens to the vocabulary. Consistent with our biomedical ablation, the experiment with $\alpha$ parameters produced similar trends, with $\alpha=0.3$ yielding the most stable and highest-performing results (see Appendix \ref{tab:islamic_alpha_ablation}). The weights for the MOSAIC-ID (Islamic Domain) and Islamic-embed-model (trained solely using a contrastive approach) are initialized with nomic-embed-text-v1\textsubscript{unsup}.

\textbf{Results}.
The MOSAIC-ID-Stage3 model achieves the highest NDCG@10 score (36.8; Table~\ref{tab:ndcg10_islamic}), surpassing both general-domain baselines and the contrastive-only variant (Islamic-embed-model). All models exhibit considerable variation in per-query scores (Figure~\ref{fig:boxplot_islamic}), reflecting the challenging nature of the dataset, but our model’s upper quartile and mean are both higher. 
Notably, the lower whisker for the MOSAIC-ID-Stage3 does not reach the minimum value of 0, whereas the lower whiskers for the general-domain models extend to 0. This indicates that our model makes fewer completely incorrect predictions (i.e., queries with NDCG@10 = 0), while the comparison models sometimes fail to retrieve any relevant results for certain queries. The upper whiskers are similar across all models, suggesting comparable best-case performance, but the reduction in low and zero scores for our model contributes to its higher overall mean NDCG@10.
This performance gap can be attributed to differences in pretraining data: while biomedical content constitutes a measurable minority of large-scale pretraining corpora \citep{Xiao2023CPackPR, Wang2022TextEB, gte2023, nomic2024}, Islamic domain texts are almost absent (typically less than 0.01\%). This negligible coverage leaves general-domain models ill-equipped to capture the linguistic and conceptual nuances of Islamic texts, making domain adaptation essential for low-resource areas \citep{pavlova-makhlouf-2024-building}. The qualitative analysis on QRCD is provided in the Appendix~\ref{sec:qualitative_analysis}.

\section{Conclusion}
We present MOSAIC, a novel approach for domain adaptation of text embedding models by jointly optimizing MLM and contrastive objectives. Unlike standard domain adaptation methods, which are typically applied at the language modeling stage or after task-specific training via data augmentation, our method leverages a model-driven approach for domain adaptation after contrastive training. We achieve robust gains in both high-resource (biomedical) and low-resource (Islamic) domains, surpassing general-domain baselines even with limited in-domain data.

\section*{Limitations}

Much of the research on domain adaptation focuses on high-resource domains such as biomedicine, where data is abundant and benchmarks are well established. In this work, we explicitly include a low-resource domain (Islamic text), recognizing both the additional challenges and the importance of extending language technologies to underrepresented settings. However, we recognize that each domain, whether high- or low-resource, can present unique characteristics and challenges that could affect the effectiveness of domain adaptation methods. Our approach relies on augmenting the existing vocabulary with domain-specific tokens, which differ across domains and may introduce challenges in tokenizer consistency; however, these effects are mitigated by the stabilizing influence of the MLM objective. As such, the generalizability of our approach may vary depending on domain-specific linguistic features, data availability, or cultural context. We encourage further research on adaptation strategies that are sensitive to the specific requirements and risks of diverse domains. 

\section*{Ethical Considerations}
Adapting models to specialized domains may amplify biases or inaccuracies present in domain-specific corpora. For example, biomedical texts may reflect publication biases or outdated medical practices, while religious texts may encode culturally specific viewpoints. In our work, we rely exclusively on publicly available and verified resources for data collection and model training; no private or proprietary data is used at any stage. Nevertheless, we acknowledge that these sources may still carry implicit biases or limitations. We encourage users of domain-adapted models to consider these factors carefully, especially when applying the models in sensitive or high-impact contexts. The models will be released under the Apache-2.0 license to ensure transparency, reproducibility, and broad accessibility. The model \texttt{nomic-embed-text-v1\textsubscript{unsup}} is licensed under Apache-2.0. All artifacts used in this study are open-source and available for research purposes.
We utilized AI assistants for debugging, optimizing LaTeX formatting, and improving grammar clarity. 

\section*{Acknowledgment}
We gratefully acknowledge the Dubai AI Campus for providing the computing infrastructure that made this work possible. We extend our appreciation to our colleagues and collaborators for their support and inspiration throughout this project.

\bibliography{anthology,custom}

\begin{thebibliography}{53}
\providecommand{\natexlab}[1]{#1}

\bibitem[{Alsentzer et~al.(2019)Alsentzer, Murphy, Boag, Weng, Jindi, Naumann, and McDermott}]{alsentzer-etal-2019-publicly}
Emily Alsentzer, John Murphy, William Boag, Wei-Hung Weng, Di~Jindi, Tristan Naumann, and Matthew McDermott. 2019.
\newblock \href {https://doi.org/10.18653/v1/W19-1909} {Publicly available clinical {BERT} embeddings}.
\newblock In \emph{Proceedings of the 2nd Clinical Natural Language Processing Workshop}, pages 72--78, Minneapolis, Minnesota, USA. Association for Computational Linguistics.

\bibitem[{Bachman et~al.(2019)Bachman, Hjelm, and Buchwalter}]{Bachman2019LearningRB}
Philip Bachman, R.~Devon Hjelm, and William Buchwalter. 2019.
\newblock \href {https://api.semanticscholar.org/CorpusID:173990164} {Learning representations by maximizing mutual information across views}.
\newblock In \emph{Neural Information Processing Systems}.

\bibitem[{Belfathi et~al.(2024)Belfathi, Gallina, Hernandez, Dufour, and Monceaux}]{Belfathi2024LanguageMA}
Anas Belfathi, Ygor Gallina, Nicolas Hernandez, Richard Dufour, and Laura Monceaux. 2024.
\newblock \href {https://api.semanticscholar.org/CorpusID:267750231} {Language model adaptation to specialized domains through selective masking based on genre and topical characteristics}.
\newblock \emph{ArXiv}, abs/2402.12036.

\bibitem[{Beltagy et~al.(2019)Beltagy, Lo, and Cohan}]{beltagy-etal-2019-scibert}
Iz~Beltagy, Kyle Lo, and Arman Cohan. 2019.
\newblock \href {https://doi.org/10.18653/v1/D19-1371} {{S}ci{BERT}: A pretrained language model for scientific text}.
\newblock In \emph{Proceedings of the 2019 Conference on Empirical Methods in Natural Language Processing and the 9th International Joint Conference on Natural Language Processing (EMNLP-IJCNLP)}, pages 3615--3620, Hong Kong, China. Association for Computational Linguistics.

\bibitem[{Caruana(1997)}]{caruana1997multitask}
Rich Caruana. 1997.
\newblock \href {https://doi.org/10.1023/A:1007379606734} {Multitask learning}.
\newblock \emph{Machine Learning}, 28(1):41--75.

\bibitem[{Cer et~al.(2017)Cer, Diab, Agirre, Lopez-Gazpio, and Specia}]{cer-etal-2017-semeval}
Daniel Cer, Mona Diab, Eneko Agirre, I{\~n}igo Lopez-Gazpio, and Lucia Specia. 2017.
\newblock \href {https://doi.org/10.18653/v1/S17-2001} {{S}em{E}val-2017 task 1: Semantic textual similarity multilingual and crosslingual focused evaluation}.
\newblock In \emph{Proceedings of the 11th International Workshop on Semantic Evaluation ({S}em{E}val-2017)}, pages 1--14, Vancouver, Canada. Association for Computational Linguistics.

\bibitem[{Chen et~al.(2020)Chen, Kornblith, Norouzi, and Hinton}]{Chen2020ASF}
Ting Chen, Simon Kornblith, Mohammad Norouzi, and Geoffrey~E. Hinton. 2020.
\newblock \href {https://api.semanticscholar.org/CorpusID:211096730} {A simple framework for contrastive learning of visual representations}.
\newblock \emph{ArXiv}, abs/2002.05709.

\bibitem[{Chi et~al.(2021)Chi, Dong, Wei, Yang, Singhal, Wang, Song, Mao, Huang, and Zhou}]{chi-etal-2021-infoxlm}
Zewen Chi, Li~Dong, Furu Wei, Nan Yang, Saksham Singhal, Wenhui Wang, Xia Song, Xian-Ling Mao, Heyan Huang, and Ming Zhou. 2021.
\newblock \href {https://doi.org/10.18653/v1/2021.naacl-main.280} {{I}nfo{XLM}: An information-theoretic framework for cross-lingual language model pre-training}.
\newblock In \emph{Proceedings of the 2021 Conference of the North American Chapter of the Association for Computational Linguistics: Human Language Technologies}, pages 3576--3588, Online. Association for Computational Linguistics.

\bibitem[{Conneau et~al.(2020)Conneau, Khandelwal, Goyal, Chaudhary, Wenzek, Guzm{\'a}n, Grave, Ott, Zettlemoyer, and Stoyanov}]{conneau-etal-2020-unsupervised}
Alexis Conneau, Kartikay Khandelwal, Naman Goyal, Vishrav Chaudhary, Guillaume Wenzek, Francisco Guzm{\'a}n, Edouard Grave, Myle Ott, Luke Zettlemoyer, and Veselin Stoyanov. 2020.
\newblock \href {https://doi.org/10.18653/v1/2020.acl-main.747} {Unsupervised cross-lingual representation learning at scale}.
\newblock In \emph{Proceedings of the 58th Annual Meeting of the Association for Computational Linguistics}, pages 8440--8451, Online. Association for Computational Linguistics.

\bibitem[{Devlin et~al.(2019)Devlin, Chang, Lee, and Toutanova}]{devlin-etal-2019-bert}
Jacob Devlin, Ming-Wei Chang, Kenton Lee, and Kristina Toutanova. 2019.
\newblock \href {https://doi.org/10.18653/v1/N19-1423} {{BERT}: Pre-training of deep bidirectional transformers for language understanding}.
\newblock In \emph{Proceedings of the 2019 Conference of the North {A}merican Chapter of the Association for Computational Linguistics: Human Language Technologies, Volume 1 (Long and Short Papers)}, pages 4171--4186, Minneapolis, Minnesota. Association for Computational Linguistics.

\bibitem[{Ethayarajh(2019)}]{ethayarajh-2019-contextual}
Kawin Ethayarajh. 2019.
\newblock \href {https://doi.org/10.18653/v1/D19-1006} {How contextual are contextualized word representations? {C}omparing the geometry of {BERT}, {ELM}o, and {GPT}-2 embeddings}.
\newblock In \emph{Proceedings of the 2019 Conference on Empirical Methods in Natural Language Processing and the 9th International Joint Conference on Natural Language Processing (EMNLP-IJCNLP)}, pages 55--65, Hong Kong, China. Association for Computational Linguistics.

\bibitem[{Gao et~al.(2021)Gao, Yao, and Chen}]{gao-etal-2021-simcse}
Tianyu Gao, Xingcheng Yao, and Danqi Chen. 2021.
\newblock \href {https://doi.org/10.18653/v1/2021.emnlp-main.552} {{S}im{CSE}: Simple contrastive learning of sentence embeddings}.
\newblock In \emph{Proceedings of the 2021 Conference on Empirical Methods in Natural Language Processing}, pages 6894--6910, Online and Punta Cana, Dominican Republic. Association for Computational Linguistics.

\bibitem[{Giorgi et~al.(2021)Giorgi, Nitski, Wang, and Bader}]{giorgi-etal-2021-declutr}
John Giorgi, Osvald Nitski, Bo~Wang, and Gary Bader. 2021.
\newblock \href {https://doi.org/10.18653/v1/2021.acl-long.72} {{D}e{CLUTR}: Deep contrastive learning for unsupervised textual representations}.
\newblock In \emph{Proceedings of the 59th Annual Meeting of the Association for Computational Linguistics and the 11th International Joint Conference on Natural Language Processing (Volume 1: Long Papers)}, pages 879--895, Online. Association for Computational Linguistics.

\bibitem[{Gu et~al.(2020{\natexlab{a}})Gu, Tinn, Cheng, Lucas, Usuyama, Liu, Naumann, Gao, and Poon}]{Gu-PubMedBERT}
Yu~Gu, Robert Tinn, Hao Cheng, Michael Lucas, Naoto Usuyama, Xiaodong Liu, Tristan Naumann, Jianfeng Gao, and Hoifung Poon. 2020{\natexlab{a}}.
\newblock \href {https://arxiv.org/abs/2007.15779} {Domain-specific language model pretraining for biomedical natural language processing}.
\newblock \emph{CoRR}, abs/2007.15779.

\bibitem[{Gu et~al.(2020{\natexlab{b}})Gu, Zhang, Wang, Liu, and Sun}]{gu-etal-2020-train}
Yuxian Gu, Zhengyan Zhang, Xiaozhi Wang, Zhiyuan Liu, and Maosong Sun. 2020{\natexlab{b}}.
\newblock \href {https://doi.org/10.18653/v1/2020.emnlp-main.566} {Train no evil: Selective masking for task-guided pre-training}.
\newblock In \emph{Proceedings of the 2020 Conference on Empirical Methods in Natural Language Processing (EMNLP)}, pages 6966--6974, Online. Association for Computational Linguistics.

\bibitem[{Hjelm et~al.(2018)Hjelm, Fedorov, Lavoie-Marchildon, Grewal, Trischler, and Bengio}]{Hjelm2018LearningDR}
R.~Devon Hjelm, Alex Fedorov, Samuel Lavoie-Marchildon, Karan Grewal, Adam Trischler, and Yoshua Bengio. 2018.
\newblock \href {https://api.semanticscholar.org/CorpusID:52055130} {Learning deep representations by mutual information estimation and maximization}.
\newblock \emph{ArXiv}, abs/1808.06670.

\bibitem[{Huang et~al.(2023)Huang, Wang, Dutta, Patel, Glava{\v{s}}, and Gurevych}]{huang-etal-2023-adasent}
Yongxin Huang, Kexin Wang, Sourav Dutta, Raj Patel, Goran Glava{\v{s}}, and Iryna Gurevych. 2023.
\newblock \href {https://doi.org/10.18653/v1/2023.emnlp-main.208} {{A}da{S}ent: Efficient domain-adapted sentence embeddings for few-shot classification}.
\newblock In \emph{Proceedings of the 2023 Conference on Empirical Methods in Natural Language Processing}, pages 3420--3434, Singapore. Association for Computational Linguistics.

\bibitem[{Kobak et~al.(2024)Kobak, He et~al.}]{kobak2024landscape}
Dmitry Kobak, Dong He, et~al. 2024.
\newblock \href {https://doi.org/10.1101/2023.04.10.536208} {The landscape of biomedical research}.
\newblock \emph{bioRxiv}.

\bibitem[{Kong et~al.(2019)Kong, de~Masson~d'Autume, Ling, Yu, Dai, and Yogatama}]{Kong2019AMI}
Lingpeng Kong, Cyprien de~Masson~d'Autume, Wang Ling, Lei Yu, Zihang Dai, and Dani Yogatama. 2019.
\newblock \href {https://api.semanticscholar.org/CorpusID:204788776} {A mutual information maximization perspective of language representation learning}.
\newblock \emph{ArXiv}, abs/1910.08350.

\bibitem[{Lee et~al.(2019)Lee, Yoon, Kim, Kim, Kim, So, and Kang}]{Lee-2019}
Jinhyuk Lee, Wonjin Yoon, Sungdong Kim, Donghyeon Kim, Sunkyu Kim, Chan~Ho So, and Jaewoo Kang. 2019.
\newblock \href {https://doi.org/10.1093/bioinformatics/btz682} {{BioBERT}: a pre-trained biomedical language representation model for biomedical text mining}.
\newblock \emph{Bioinformatics}.

\bibitem[{Li et~al.(2020)Li, Zhou, He, Wang, Yang, and Li}]{li-etal-2020-sentence}
Bohan Li, Hao Zhou, Junxian He, Mingxuan Wang, Yiming Yang, and Lei Li. 2020.
\newblock \href {https://doi.org/10.18653/v1/2020.emnlp-main.733} {On the sentence embeddings from pre-trained language models}.
\newblock In \emph{Proceedings of the 2020 Conference on Empirical Methods in Natural Language Processing (EMNLP)}, pages 9119--9130, Online. Association for Computational Linguistics.

\bibitem[{Li et~al.(2023)Li, Zhang, Pan, Li, Ding, Wu, Wang, Zhu, and Huang}]{gte2023}
Yining Li, Yuhui Zhang, Xiaoman Pan, Yinan Li, Ning Ding, Wei Wu, Yujing Wang, Xiaoyan Zhu, and Minlie Huang. 2023.
\newblock \href {https://arxiv.org/abs/2308.03281} {Towards general text embeddings with multi-stage contrastive learning}.
\newblock \emph{arXiv preprint arXiv:2308.03281}.

\bibitem[{Liu et~al.(2021)Liu, Vuli{\'c}, Korhonen, and Collier}]{liu-etal-2021-fast}
Fangyu Liu, Ivan Vuli{\'c}, Anna Korhonen, and Nigel Collier. 2021.
\newblock \href {https://doi.org/10.18653/v1/2021.emnlp-main.109} {Fast, effective, and self-supervised: Transforming masked language models into universal lexical and sentence encoders}.
\newblock In \emph{Proceedings of the 2021 Conference on Empirical Methods in Natural Language Processing}, pages 1442--1459, Online and Punta Cana, Dominican Republic. Association for Computational Linguistics.

\bibitem[{Liu et~al.(2020)Liu, Ott, Goyal, Du, Joshi, Chen, Levy, Lewis, Zettlemoyer, and Stoyanov}]{liu-etal-2020-roberta}
Yinhan Liu, Myle Ott, Naman Goyal, Jingfei Du, Mandar Joshi, Danqi Chen, Omer Levy, Mike Lewis, Luke Zettlemoyer, and Veselin Stoyanov. 2020.
\newblock \href {https://aclanthology.org/2020.acl-main.103} {{R}o{BERT}a: A robustly optimized {BERT} pretraining approach}.
\newblock In \emph{Proceedings of the 58th Annual Meeting of the Association for Computational Linguistics}, pages 1--10, Online. Association for Computational Linguistics.

\bibitem[{Malhas and Elsayed(2020)}]{malhas2020ayatec}
Rana Malhas and Tamer Elsayed. 2020.
\newblock Ayatec: building a reusable verse-based test collection for arabic question answering on the holy qur’an.
\newblock \emph{ACM Transactions on Asian and Low-Resource Language Information Processing (TALLIP)}, 19(6):1--21.

\bibitem[{Meng et~al.(2021)Meng, Xiong, Bajaj, Tiwary, Bennett, Han, and Song}]{meng2021coco}
Yu~Meng, Chenyan Xiong, Payal Bajaj, Saurabh Tiwary, Paul~N. Bennett, Jiawei Han, and Xia Song. 2021.
\newblock \href {https://api.semanticscholar.org/CorpusID:231942621} {Coco-lm: Correcting and contrasting text sequences for language model pretraining}.
\newblock In \emph{Neural Information Processing Systems}.

\bibitem[{Merrick et~al.(2024)Merrick, Beeler, Lewis, Girdhar, Hall, Chen, Thickstun, and Devlin}]{arctic2024}
Logan Merrick, Cameron Beeler, Mike Lewis, Rohit Girdhar, David Hall, Xilun Chen, John Thickstun, and Jacob Devlin. 2024.
\newblock \href {https://arxiv.org/abs/2405.05374} {Arctic-embed: Scalable, efficient, and accurate text embedding models}.
\newblock \emph{arXiv preprint arXiv:2405.05374}.

\bibitem[{Muennighoff et~al.(2022)Muennighoff, Tazi, Magne, and Reimers}]{MTEB}
Niklas Muennighoff, Nouamane Tazi, Loic Magne, and Nils Reimers. 2022.
\newblock \href {https://api.semanticscholar.org/CorpusID:252907685} {Mteb: Massive text embedding benchmark}.
\newblock In \emph{Conference of the European Chapter of the Association for Computational Linguistics}.

\bibitem[{Nussbaum et~al.(2024)Nussbaum, Morris, Duderstadt, and Mulyar}]{nomic2024}
Zach Nussbaum, John~X. Morris, Brandon Duderstadt, and Andriy Mulyar. 2024.
\newblock \href {https://arxiv.org/abs/2402.01613} {Nomic embed: Training a reproducible long context text embedder}.
\newblock \emph{arXiv preprint arXiv:2402.01613}.

\bibitem[{Pavlova(2023)}]{pavlova-2023-leveraging}
Vera Pavlova. 2023.
\newblock \href {https://doi.org/10.18653/v1/2023.arabicnlp-1.7} {Leveraging domain adaptation and data augmentation to improve qur{'}anic {IR} in {E}nglish and {A}rabic}.
\newblock In \emph{Proceedings of ArabicNLP 2023}, pages 76--88, Singapore (Hybrid). Association for Computational Linguistics.

\bibitem[{Pavlova(2025)}]{pavlova-2025-multi}
Vera Pavlova. 2025.
\newblock \href {https://aclanthology.org/2025.clrel-1.4/} {Multi-stage training of bilingual islamic {LLM} for neural passage retrieval}.
\newblock In \emph{Proceedings of the New Horizons in Computational Linguistics for Religious Texts}, pages 42--52, Abu Dhabi, UAE. Association for Computational Linguistics.

\bibitem[{Pavlova and Makhlouf(2023)}]{pavlova-makhlouf-2023-bioptimus}
Vera Pavlova and Mohammed Makhlouf. 2023.
\newblock \href {https://doi.org/10.18653/v1/2023.bionlp-1.31} {{BIO}ptimus: Pre-training an optimal biomedical language model with curriculum learning for named entity recognition}.
\newblock In \emph{The 22nd Workshop on Biomedical Natural Language Processing and BioNLP Shared Tasks}, pages 337--349, Toronto, Canada. Association for Computational Linguistics.

\bibitem[{Pavlova and Makhlouf(2024)}]{pavlova-makhlouf-2024-building}
Vera Pavlova and Mohammed Makhlouf. 2024.
\newblock \href {https://doi.org/10.18653/v1/2024.emnlp-industry.73} {Building an efficient multilingual non-profit {IR} system for the islamic domain leveraging multiprocessing design in rust}.
\newblock In \emph{Proceedings of the 2024 Conference on Empirical Methods in Natural Language Processing: Industry Track}, pages 981--990, Miami, Florida, US. Association for Computational Linguistics.

\bibitem[{Pavlova and Makhlouf(2025)}]{pavlova-makhlouf-2025-efficient}
Vera Pavlova and Mohammed Makhlouf. 2025.
\newblock \href {https://doi.org/10.18653/v1/2025.emnlp-industry.86} {Efficient and versatile model for multilingual information retrieval of islamic text: Development and deployment in real-world scenarios}.
\newblock In \emph{Proceedings of the 2025 Conference on Empirical Methods in Natural Language Processing: Industry Track}, pages 1239--1249, Suzhou (China). Association for Computational Linguistics.

\bibitem[{Poerner et~al.(2020)Poerner, Waltinger, and Sch{\"u}tze}]{Poerner-etal-2020-inexpensive}
Nina Poerner, Ulli Waltinger, and Hinrich Sch{\"u}tze. 2020.
\newblock \href {https://doi.org/10.18653/v1/2020.findings-emnlp.134} {Inexpensive domain adaptation of pretrained language models: Case studies on biomedical {NER} and covid-19 {QA}}.
\newblock In \emph{Findings of the Association for Computational Linguistics: EMNLP 2020}, pages 1482--1490, Online. Association for Computational Linguistics.

\bibitem[{Reimers and Gurevych(2019)}]{reimers-gurevych-2019-sentence}
Nils Reimers and Iryna Gurevych. 2019.
\newblock \href {https://doi.org/10.18653/v1/D19-1410} {Sentence-{BERT}: Sentence embeddings using {S}iamese {BERT}-networks}.
\newblock In \emph{Proceedings of the 2019 Conference on Empirical Methods in Natural Language Processing and the 9th International Joint Conference on Natural Language Processing (EMNLP-IJCNLP)}, pages 3982--3992, Hong Kong, China. Association for Computational Linguistics.

\bibitem[{Sachidananda et~al.(2021)Sachidananda, Kessler, and Lai}]{sachidananda-etal-2021-efficient}
Vin Sachidananda, Jason Kessler, and Yi-An Lai. 2021.
\newblock \href {https://doi.org/10.18653/v1/2021.sustainlp-1.16} {Efficient domain adaptation of language models via adaptive tokenization}.
\newblock In \emph{Proceedings of the Second Workshop on Simple and Efficient Natural Language Processing}, pages 155--165, Virtual. Association for Computational Linguistics.

\bibitem[{Sadeq et~al.(2022)Sadeq, Xu, and McAuley}]{sadeq-etal-2022-informask}
Nafis Sadeq, Canwen Xu, and Julian McAuley. 2022.
\newblock \href {https://doi.org/10.18653/v1/2022.emnlp-main.395} {{I}nfor{M}ask: Unsupervised informative masking for language model pretraining}.
\newblock In \emph{Proceedings of the 2022 Conference on Empirical Methods in Natural Language Processing}, pages 5866--5878, Abu Dhabi, United Arab Emirates. Association for Computational Linguistics.

\bibitem[{Sanh et~al.(2019)Sanh, Debut, Chaumond, and Wolf}]{Sanh-distillation}
Victor Sanh, Lysandre Debut, Julien Chaumond, and Thomas Wolf. 2019.
\newblock \href {https://arxiv.org/abs/1910.01108} {Distilbert, a distilled version of {BERT:} smaller, faster, cheaper and lighter}.
\newblock \emph{CoRR}, abs/1910.01108.

\bibitem[{Schopf et~al.(2023)Schopf, Schneider, and Matthes}]{schopf-etal-2023-efficient}
Tim Schopf, Dennis~N. Schneider, and Florian Matthes. 2023.
\newblock \href {https://aclanthology.org/2023.ranlp-1.112} {Efficient domain adaptation of sentence embeddings using adapters}.
\newblock In \emph{Proceedings of the 14th International Conference on Recent Advances in Natural Language Processing}, pages 1046--1053, Varna, Bulgaria. INCOMA Ltd., Shoumen, Bulgaria.

\bibitem[{Sogancioglu et~al.(2017)Sogancioglu, Ozgur, and Ozturk}]{biosses}
Gizem Sogancioglu, Arzucan Ozgur, and Hakime Ozturk. 2017.
\newblock \href {https://doi.org/10.1093/bioinformatics/btx231} {Biosses: a semantic sentence similarity estimation system for the biomedical domain}.
\newblock \emph{Bioinformatics}, 33(14):i49--i58.

\bibitem[{Thakur et~al.(2021)Thakur, Reimers, Daxenberger, and Gurevych}]{thakur-etal-2021-augmented}
Nandan Thakur, Nils Reimers, Johannes Daxenberger, and Iryna Gurevych. 2021.
\newblock \href {https://doi.org/10.18653/v1/2021.naacl-main.28} {Augmented {SBERT}: Data augmentation method for improving bi-encoders for pairwise sentence scoring tasks}.
\newblock In \emph{Proceedings of the 2021 Conference of the North American Chapter of the Association for Computational Linguistics: Human Language Technologies}, pages 296--310, Online. Association for Computational Linguistics.

\bibitem[{Tsatsaronis et~al.(2015)Tsatsaronis, Balikas, Malakasiotis, Partalas, Zschunke, Alvers, Weissenborn, Krithara, Petersen, Hakenberg et~al.}]{tsatsaronis2015bioasq}
George Tsatsaronis, Georgios Balikas, Prodromos Malakasiotis, Ioannis Partalas, Matthias Zschunke, Michael~R Alvers, Dirk Weissenborn, Anastasia Krithara, Jens Petersen, J{\"o}rg Hakenberg, et~al. 2015.
\newblock An overview of the bioasq large-scale biomedical semantic indexing and question answering competition.
\newblock In \emph{BMC bioinformatics}, volume~16, page 138. BioMed Central.

\bibitem[{van~den Oord et~al.(2018)van~den Oord, Li, and Vinyals}]{Oord-Contrastive}
A{\"{a}}ron van~den Oord, Yazhe Li, and Oriol Vinyals. 2018.
\newblock \href {https://arxiv.org/abs/1807.03748} {Representation learning with contrastive predictive coding}.
\newblock \emph{CoRR}, abs/1807.03748.

\bibitem[{Vaswani et~al.(2017)Vaswani, Shazeer, Parmar, Uszkoreit, Jones, Gomez, Kaiser, and Polosukhin}]{Vaswani-transformers}
Ashish Vaswani, Noam Shazeer, Niki Parmar, Jakob Uszkoreit, Llion Jones, Aidan~N. Gomez, Lukasz Kaiser, and Illia Polosukhin. 2017.
\newblock \href {https://arxiv.org/abs/1706.03762} {Attention is all you need}.
\newblock \emph{CoRR}, abs/1706.03762.

\bibitem[{Voorhees et~al.(2021)Voorhees, Alam, Bedrick, Demner-Fushman, Hersh, Roberts, and Wang}]{voorhees2021trec}
Ellen Voorhees, Tasmeer Alam, Steven Bedrick, Dina Demner-Fushman, William Hersh, Kirk Roberts, and Lucy~Lu Wang. 2021.
\newblock \href {https://doi.org/10.1093/jamia/ocab047} {Trec-covid: Constructing a pandemic information retrieval test collection}.
\newblock \emph{Journal of the American Medical Informatics Association}, 28(4):766--773.

\bibitem[{Wang et~al.(2018)Wang, Singh, Michael, Hill, Levy, and Bowman}]{wang-etal-2018-glue}
Alex Wang, Amanpreet Singh, Julian Michael, Felix Hill, Omer Levy, and Samuel Bowman. 2018.
\newblock \href {https://doi.org/10.18653/v1/W18-5446} {{GLUE}: A multi-task benchmark and analysis platform for natural language understanding}.
\newblock In \emph{Proceedings of the 2018 {EMNLP} Workshop {B}lackbox{NLP}: Analyzing and Interpreting Neural Networks for {NLP}}, pages 353--355, Brussels, Belgium. Association for Computational Linguistics.

\bibitem[{Wang et~al.(2021)Wang, Reimers, and Gurevych}]{wang-etal-2021-tsdae-using}
Kexin Wang, Nils Reimers, and Iryna Gurevych. 2021.
\newblock \href {https://doi.org/10.18653/v1/2021.findings-emnlp.59} {{TSDAE}: Using transformer-based sequential denoising auto-encoderfor unsupervised sentence embedding learning}.
\newblock In \emph{Findings of the Association for Computational Linguistics: EMNLP 2021}, pages 671--688, Punta Cana, Dominican Republic. Association for Computational Linguistics.

\bibitem[{Wang et~al.(2022{\natexlab{a}})Wang, Thakur, Reimers, and Gurevych}]{wang-etal-2022-gpl}
Kexin Wang, Nandan Thakur, Nils Reimers, and Iryna Gurevych. 2022{\natexlab{a}}.
\newblock \href {https://doi.org/10.18653/v1/2022.naacl-main.168} {{GPL}: Generative pseudo labeling for unsupervised domain adaptation of dense retrieval}.
\newblock In \emph{Proceedings of the 2022 Conference of the North American Chapter of the Association for Computational Linguistics: Human Language Technologies}, pages 2345--2360, Seattle, United States. Association for Computational Linguistics.

\bibitem[{Wang et~al.(2022{\natexlab{b}})Wang, Yang, Huang, Jiao, Yang, Jiang, Majumder, and Wei}]{Wang2022TextEB}
Liang Wang, Nan Yang, Xiaolong Huang, Binxing Jiao, Linjun Yang, Daxin Jiang, Rangan Majumder, and Furu Wei. 2022{\natexlab{b}}.
\newblock \href {https://api.semanticscholar.org/CorpusID:254366618} {Text embeddings by weakly-supervised contrastive pre-training}.
\newblock \emph{ArXiv}, abs/2212.03533.

\bibitem[{Wu et~al.(2022)Wu, Gao, Lin, Han, Wang, and Hu}]{wu-etal-2022-infocse}
Xing Wu, Chaochen Gao, Zijia Lin, Jizhong Han, Zhongyuan Wang, and Songlin Hu. 2022.
\newblock \href {https://doi.org/10.18653/v1/2022.findings-emnlp.223} {{I}nfo{CSE}: Information-aggregated contrastive learning of sentence embeddings}.
\newblock In \emph{Findings of the Association for Computational Linguistics: EMNLP 2022}, pages 3060--3070, Abu Dhabi, United Arab Emirates. Association for Computational Linguistics.

\bibitem[{Wu et~al.(2020)Wu, Wang, Gu, Khabsa, Sun, and Ma}]{wu2020clear}
Zhuofeng Wu, Sinong Wang, Jiatao Gu, Madian Khabsa, Fei Sun, and Hao Ma. 2020.
\newblock \href {https://arxiv.org/abs/2012.15466} {Clear: Contrastive learning for sentence representation}.
\newblock \emph{arXiv preprint arXiv:2012.15466}.

\bibitem[{Xiao et~al.(2023)Xiao, Liu, Zhang, Muennighoff, Lian, and yun Nie}]{Xiao2023CPackPR}
Shitao Xiao, Zheng Liu, Peitian Zhang, Niklas Muennighoff, Defu Lian, and Jian yun Nie. 2023.
\newblock \href {https://api.semanticscholar.org/CorpusID:271114619} {C-pack: Packed resources for general chinese embeddings}.
\newblock \emph{Proceedings of the 47th International ACM SIGIR Conference on Research and Development in Information Retrieval}.

\end{thebibliography}

\clearpage

\appendix

\section{Consistency-based Filtering Procedure}
\label{sec:false_positive}
To further ensure data quality and minimize false positive pairs, we employed a semantic filtering procedure using the \texttt{gte-base} model. Specifically, we first sampled up to 1 million candidate query–document pairs from the initial dataset. Each query and document was independently encoded into dense vector representations using the \texttt{gte-base} text embedding model.

Next, we constructed a FAISS index from all document embeddings to enable efficient similarity search. For each query embedding, we retrieved the top-$k$ most similar document embeddings from the index, based on cosine similarity. If the original paired document $d_i$ was not found among the top-$k$ retrieved documents for its corresponding query $q_i$, we discarded the pair $(q_i, d_i)$. This filtering step ensures that only pairs with strong semantic alignment—according to the embedding model—are retained for further training.

The intuition behind this approach is to eliminate weakly related or noisy pairs that may have been erroneously grouped together in the initial data extraction. By keeping only those pairs where the document is highly ranked for its query, we improve the quality and relevance of training examples, leading to better domain adaptation during model training.

\section{Curating Passages for Training the Islamic Domain Model}
\label{sec:islamic_passages}
Dense retrieval models often experience performance degradation when applied to new domains, emphasizing the value of training on in-domain data. The scarcity of such data is typically addressed through augmentation techniques like synthetic data generation, paraphrasing, pair recombination, round-trip translation, or denoising autoencoders. However, these approaches risk altering the original semantics, which is especially problematic for sensitive religious and heritage texts.
To overcome this, we utilize Tafseer Ibn Kathir, a classical and authoritative Qur’anic exegesis rich in verse commentary and inter-verse references. This resource enables natural and semantically meaningful augmentation of training data by explicitly linking related verses.

Pair Extraction.
Let $C_{t}$ denote the collection of Tafseer texts by Ibn Kathir. We extract all verse pairs $V_{t} = {(v_q, v_p)}$ referenced in $C_{t}$, resulting in approximately 11,000 candidate pairs.

Filtering.
Not all extracted pairs represent strong semantic correlations suitable for retrieval training, due to indirect or implicit relationships. To select high-quality positive pairs, we score each candidate $(v_q, v_p)$ using the \texttt{gte-base} model to obtain similarity scores $s = \texttt{gte-base}(v_q, v_p)$. Pairs scoring below a predefined threshold are removed, yielding a filtered set $V_f$ of 7,587 robust positive pairs for training.

\begin{table}[t]
\centering
\small
\resizebox{1.0\columnwidth}{!}{%
\begin{tabular}{lc}
\hline
\textbf{Model} & \textbf{NDCG@10} \\
\hline
MOSAIC-ID-Stage3 ($\alpha=0.1$) & 30.750 \\
MOSAIC-ID-Stage3 ($\alpha=0.2$) & 34.622 \\
MOSAIC-ID-Stage3 ($\alpha=0.3$) & \textbf{36.809} \\
MOSAIC-ID-Stage3 ($\alpha=0.4$) & 35.251 \\
MOSAIC-ID-Stage3 ($\alpha=0.5$) & 33.049 \\
\hline
\end{tabular}
}
\caption{Ablation of the $\alpha$ hyperparameter in MOSAIC-ID-Stage3 on the Islamic dataset.}
\label{tab:islamic_alpha_ablation}
\end{table}

\begin{figure}[t]
\centering
  \includegraphics[width=\columnwidth]{{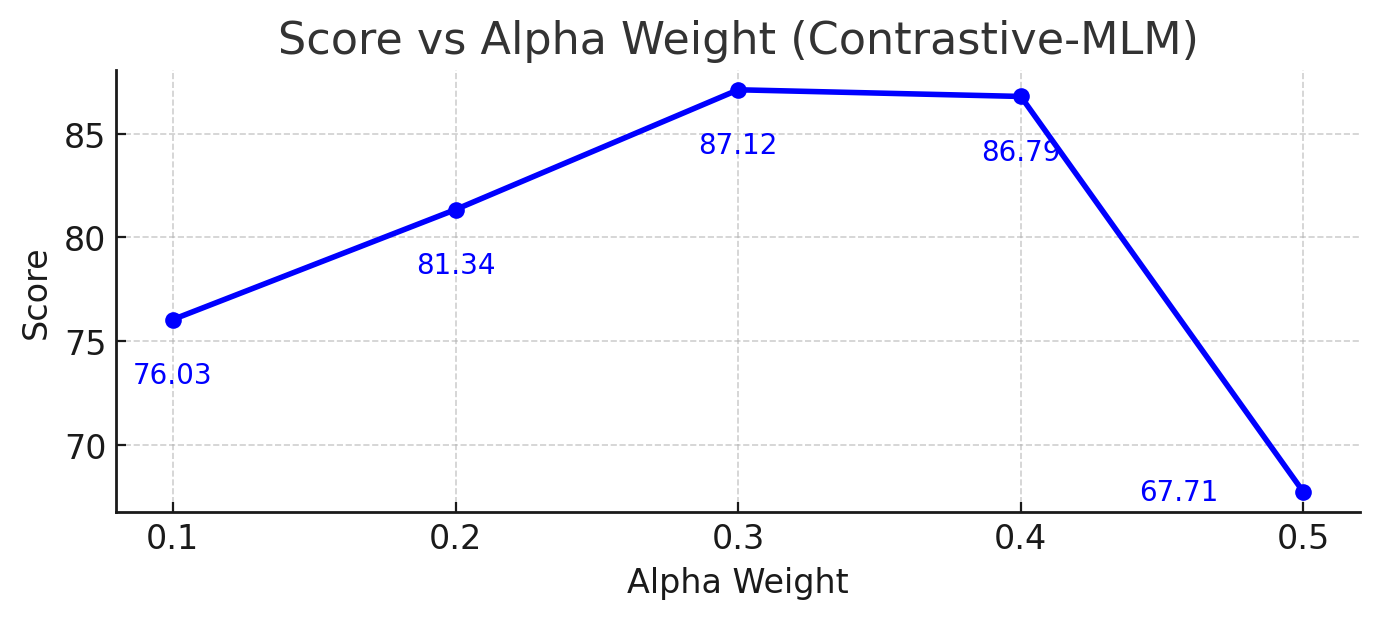}}
  \caption{Effect of alpha weight on the performance in the 2nd stage of Contrastive+MLM training on BIOSSES.}
  \label{fig:bjmc_alpha_plot}
\end{figure}

\section{Training Hyperparameters}
\label{sec:hyperparameters}
\begin{table}[h]
\begin{adjustbox}{width=\columnwidth,center}
\begin{tabular}{ccc}
\hline
\textbf{Computing Infrastructure}    & \multicolumn{2}{c}{1x H100 (80 GB)}    \\ \hline
\multicolumn{2}{c}{\textbf{Hyperparameter}} & \textbf{Assignment}                     \\ \hline
\multicolumn{2}{c}{number of epochs}        & 1-5 \\ \hline
\multicolumn{2}{c}{batch size}              & 128                             \\ \hline
\multicolumn{2}{c}{query sequence length}   & 64-256                        \\ \hline
\multicolumn{2}{c}{maximum learning rate}   & 0.0005                        \\ \hline
\multicolumn{2}{c}{learning rate optimizer} & Adam                                    \\ \hline
\multicolumn{2}{c}{learning rate scheduler} & None or Warmup linear                   \\ \hline
\multicolumn{2}{c}{Weight decay}            & 0.01                                    \\ \hline
\multicolumn{2}{c}{Warmup proportion}       & 0.06                                    \\ \hline
\multicolumn{2}{c}{learning rate decay}     & linear                                  \\ \hline
\end{tabular}
\end{adjustbox}
\caption{Hyperparameters for training and finetuning text embedding models.}
\label{tab:appendix-table-a}
\end{table}

\section{Computational Efficiency}

\begin{table*}[t]
\begin{adjustbox}{width=\textwidth,center}
\begin{tabular}{c|c|c}
\hline
\textbf{Method} & \textbf{Relative Compute Cost} & \textbf{Notes} \\
\hline
Baseline: Contrastive FT & 1.0x & Supervised dual-encoder finetuning \\
\hline
MOSAIC -- Stage 1 (Tokenizer extension) & 0x & No GPU cost \\
\hline
MOSAIC -- Stage 2 (Domain tokens only) & ~2.0x & MLM applied only to new domain tokens \\
\hline
MOSAIC -- Stage 2 (All tokens) & ~2.5x & Upper bound: MLM applies to entire vocabulary \\
\hline
MOSAIC -- Total & ~3.0x & Stage 1 + Stage 2 (domain tokens) + CT finetuning \\
\hline
Full Domain MLM (DAPT) & 10--15x & Typical costs reported in prior work \\
\hline
\end{tabular}
\end{adjustbox}
\caption{Relative compute cost comparison between MOSAIC stages and common domain adaptation baselines.}
\label{tab:compute_cost_comparison}
\end{table*}

As summarized in (Table~\ref{tab:compute_cost_comparison}), the computational footprint of MOSAIC remains modest. The second stage introduces an overhead of approximately 2× relative to standard contrastive finetuning, while the full three-stage pipeline remains substantially cheaper than conventional domain-adaptive pretraining (DAPT). Notably, the naïve all-tokens MLM variant incurs higher computational cost, which highlights the efficiency benefits of restricting masked language modeling to newly introduced domain tokens. Overall, these results show that MOSAIC can achieve meaningful performance gains in specialized domains without requiring prohibitive computational resources.

\section{Qualitative Analysis}
\label{sec:qualitative_analysis}
For the qualitative analysis (Table~\ref{tab:appendix_biosses_examples}), we present examples from the BIOSSES dataset. In the first two cases, the MOSAIC-Stage2 model produces similarity predictions that are close to the gold scores, indicating well-calibrated semantic judgments. In contrast, the All-Tokens MLM model consistently overshoots, assigning higher similarity scores than warranted. In the remaining examples, MOSAIC-Stage2 also overestimates similarity, but to a noticeably smaller extent than the All-Tokens MLM baseline. Given the high density of biomedical terminology in BIOSSES, the All-Tokens model appears biased toward inflated similarity estimates. This behavior suggests that, during the joint MLM+contrastive stage, the All-Tokens approach does not sufficiently emphasize domain-specific token distinctions, leading to weaker fine-grained semantic discrimination across biomedical topics.

\begin{table*}[t]
\begin{adjustbox}{width=\textwidth,center}
\small
\begin{tabular}{c|p{0.33\linewidth}|p{0.33\linewidth}|c|c|c}
\hline
\textbf{ID} & \textbf{Sentence 1} & \textbf{Sentence 2} & \textbf{Gold} & \textbf{MOSAIC-Stage2} & \textbf{All-Token MLM} \\
 &  &  &  & ($\alpha=0.3$) & ($\alpha=0.001$) \\
\hline
80 &
BAF53 and $\beta$-actin subunits have been implicated in mammalian SWI/SNF-regulated binding to the chromatin/nuclear matrix. &
$\beta$-actin and actin-related proteins have been found in various ATP-dependent chromatin remodeling complexes. &
3 &
3.96 &
4.38 \\
\hline
22 &
Furthermore, a recent study demonstrated the mechanism specifying myeloid expression of miR-223 and proposed a unique minicircuitry comprised of miR-223 and transcription factors NFI-A and C/EBP$\alpha$. &
miR-223 regulates granulopoiesis by a feedback mechanism and is competitively modulated by the transcription factors NFI-A and C/EBP$\alpha$. &
3.4 &
4.08 &
4.54 \\
\hline
92 &
The cyclin-dependent kinase inhibitor roscovitine has been reported to down-regulate the anti-apoptotic protein Mcl-1. &
Recent work in model systems and acute myelogenous leukemia suggests that MCL-1 expression is a key determinant of resistance to ABT-737. &
2 &
3.42 &
3.95 \\
\hline
90 &
Necroptosis is also part of host defense against virus infection. &
Necrotic cell death was augmented when caspase activity was compromised by viral or chemical inhibitors. &
2 &
3.87 &
4.21 \\
\hline
\end{tabular}
\end{adjustbox}
\caption{Qualitative analysis examples from the BIOSSES dataset comparing MOSAIC-Stage2 with an all-token MLM baseline. Predicted scores indicate semantic similarity.}
\label{tab:appendix_biosses_examples}
\end{table*}

For the qualitative analysis in the Islamic domain, we focus on the retrieval task from the QRCD benchmark \citep{malhas2020ayatec}. Many queries in QRCD are genuinely challenging even for human readers, as they require recognizing metaphorical expressions and implicit semantic correspondences rather than relying on surface lexical overlap. To illustrate the effect of domain adaptation, we analyze two particularly challenging queries and compare the top three retrieved passages across models. We present one case in which the proposed MOSAIC-Stage3 model succeeds while baseline models fail, and a second case in which MOSAIC-Stage3 also fails, highlighting limitations that remain shared across all systems.

For \textbf{Query 308} (\textit{With what did Allah compare this life to?}), MOSAIC-Stage3 successfully retrieves two of the three gold passages within its top-3 results, whereas the baseline models—including those shown in the table—fail to retrieve any gold passage within their top-10 rankings (we report only the top-3 for brevity). This query is particularly challenging because, although phrased broadly, it requires identifying specific metaphorical analogies used in the Qur’an to describe the transient nature of worldly life. The baseline models tend to gravitate toward generally related thematic verses (e.g., monotheism, devotion, or revelation), rather than capturing the intended comparative semantic structure. In contrast, MOSAIC-Stage3 correctly prioritizes passages that employ the canonical imagery of rain-fed vegetation to convey the temporary and illusory character of worldly existence.

For \textbf{Query 350} (\textit{What are the verses in which healing is mentioned?}), none of the models retrieve any gold passage within their top-10 ranked results (again, only the top-3 are shown). All retrieved passages are thematically distant from the target concept of healing, indicating a shared difficulty in linking the query to the appropriate Qur’anic contexts. Instead, the models gravitate toward broadly devotional or theological verses with only superficial topical or lexical relevance. This failure case illustrates that, even with domain adaptation, QRCD contains queries that demand highly specific semantic grounding—including recognition of rare thematic cues and uncommon lexical expressions—which remains beyond the current capabilities of all evaluated models.

\begin{table*}[t]
\begin{adjustbox}{width=\textwidth,center}
\small
\begin{tabular}{p{0.18\linewidth}|p{0.78\linewidth}}
\hline
\textbf{Model} & \textbf{Top Retrieved Passages (abridged)} \\
\hline
\multicolumn{2}{l}{\textbf{Query 308:} With what did Allah compare this life to?} \\
\hline
MOSAIC (Stage 3) &
(1) (doc 600) ``The worldly life resembles the plants that blossom by the help of the water which God sends from the sky. After a short time all of them fade away and the winds scatter them...'' \textbf{[GOLD]}; 
(2) (doc 428) ``The example of the worldly life is like the water sent down from the sky which becomes mixed with the earth's produce that people and cattle consume. When the land becomes fertile and pleasant, people think that they have control over it...'' \textbf{[GOLD]}; 
(3) (doc 1250) ``Those whose good deeds will weigh heavier will live a pleasant life...'' \\
\hline
Islamic-embed-model &
(1) (doc 1263) ``Say, `He is the only God; there is none equal to Him.' ''; 
(2) (doc 585) ``If all humans and jinn gathered to bring the like of this Qur’an...''; 
(3) (doc 1264) ``Say, `I seek protection from the Lord of the Dawn...' '' \\
\hline
nomic-embed-text-v1 &
(1) (doc 987) ``Have you seen the one who has chosen his desires as his lord? … `The only life is this worldly life.' ''; 
(2) (doc 94) ``Have you heard about the one who argued with Abraham about his Lord...''; 
(3) (doc 908) ``We have given all kinds of examples for mankind...'' \\
\hline
\multicolumn{2}{l}{\textbf{Query 350:} What are the verses in which healing is mentioned?} \\
\hline
MOSAIC (Stage 3) &
(1) (doc 1237) ``After every difficulty there is relief; when you are free, strive and be devoted to your Lord.''; 
(2) (doc 166) ``God wants to guide you and forgive you; He wants to relieve your burden, mankind was created weak.''; 
(3) (doc 1172) ``This chapter is a reminder; those who seek guidance will find mercy from God.'' \\
\hline
nomic-embed-text-v1 &
(1) (doc 801) ``These are the verses of wisdom, a guidance and mercy for the righteous.''; 
(2) (doc 477) ``These are the illustrious verses revealed in Arabic so you may understand.''; 
(3) (doc 938) ``A revelation from the Merciful God; verses fully expounded, with glad news and warnings.'' \\
\hline
Islamic-embed &
(1) (doc 1265) ``Say: I seek protection from the Lord of mankind, against the evil of temptations.''; 
(2) (doc 2) ``The path of those whom You blessed, not those astray.''; 
(3) (doc 1) ``You alone we worship, and from You alone we seek help; guide us to the right path.'' \\
\hline
\end{tabular}
\end{adjustbox}
\caption{Qualitative analysis examples from QRCD.}
\label{tab:appendix_islamic_retrieval_examples}
\end{table*}

\end{document}